# Benchmark for Structured Retrieval-Augmented Generation in Perovskite Photovoltaics

Qiwei Han[*1,2], Chi Zhou[2] and Gaoxiang Yuan[3]

[1]Department of Chemistry, Duke University, Durham, NC 27708, USA

[2]Department of Computer Science, Georgia Institute of Technology, Atlanta, GA 30332, USA

[3]Department of Electrical and Computer Engineering, Tandon School of Engineering, New York University, Brooklyn, NY 11201, USA

[*]Corresponding author.

**Abstract**

While retrieval-augmented language models increasingly serve as literature assistants in the physical sciences, existing benchmarks evaluate them primarily on qualitative text overlap or through model-based scoring rather than against verified experimental ground truth. In experimental materials disciplines such as perovskite photovoltaics, however, the most critical inquiries are inherently quantitative and aggregate measurements across dozens or hundreds of papers. Here, we present a quantitative question-answering benchmark where all 755 ground-truth answers are computed directly from 45,441 measured device records (40,889 human-curated records through 2021 and 4,552 records extracted from 2022–2026 literature). Benchmarking six retrieval architectures over 158,274 passages from 25,662 papers using an open 14B model on a single 24 GB GPU, standard passage retrieval tops out at 16–23% accuracy (compared to 1.6–6.5% zero-shot). In contrast, equipping the language model with read-only SQL execution over the structured device table elevates accuracy to 51.5% on historical and 50.8% on prospective literature—a 33.0 $\pm$ 1.8 percentage-point advantage over hybrid text retrieval. Aggregate queries spanning extrema, distributions, and filtered record counts prove virtually unsolvable via text retrieval alone, demonstrating that systematic table extraction is an operational prerequisite for literature-scale quantitative inquiry.

## 1 Introduction

Large language models are finding rapid adoption across chemistry and materials science for paper summarisation, property extraction, and literature-grounded question answering [1–3]. Retrieval-augmented generation (RAG) has become the primary strategy for anchoring responses to published evidence and mitigating hallucinations [4, 5]. Perovskite solar cells present an ideal, highly demanding domain for evaluating these systems: the field has generated tens of thousands of experimental papers since 2013, supported by an established community database of over 40,000 curated devices [6], and specialized assistants integrating retrieval with domain fine-tuning and knowledge graphs have begun to emerge [7].

Evaluating such literature assistants remains fundamentally problematic when benchmarks rely on n-gram overlap or LLM-as-a-judge rubrics. Fluent prose often masks critical retrieval failures, particularly when an inquiry demands synthesizing hard numbers from multiple papers. A materials researcher rarely asks questions contained within a single abstract; practical questions routinely ask which papers report a composition exceeding a target efficiency, what the distribution of open-circuit voltages is for a given contact stack, or how device parameters have shifted across publication cohorts. Answering these queries exposes a fundamental dual failure mode in existing literature architectures. On one hand, standard chunk-based text retrieval, which returns a handful of localized passages, cannot assemble evidence dispersed across dozens of studies. On the other hand, unassisted text-to-SQL fails because relational schemas lack the qualitative context of research articles and cannot bridge informal scientific vocabulary (such as resolving conceptual terms like "SAM hole contacts" or "$NiO_x$" into specific chemical strings like `Me-4PACz` or `NiO-np`) without domain semantic grounding [8]. Furthermore, evaluating models solely on historical literature risks rewarding parametric memorization rather than active retrieval [5].

Here, we tackle these dual challenges by introducing a quantitative question-answering benchmark where ground truth is deterministically computed, and systematically evaluating a hybrid retrieval architecture designed to bridge the gap between unstructured prose and structured tables. The benchmark's 755 questions are generated from templates over a relational database of 45,441 verified experimental device records, partitioning evenly between historical literature (40,889 human-curated records through 2021 [6]) and prospective literature (4,552 records extracted from 2022–2026 publications). The tasks span single-paper lookups, candidate lists, maxima, record counts, distributions, pairwise comparisons, and temporal trends. To address the limitations of both text search and unassisted database querying, we implement a hybrid structured RAG architecture: the model receives retrieved text passages for provenance and context, the relational schema, and a domain data dictionary, enabling multi-turn autonomous SQL execution before generating its final answer. All experiments evaluate an open 14-billion-parameter model [9] served on a single 24 GB desktop GPU [10].

Our results demonstrate that this hybrid structured approach directly resolves the bottlenecks that paralyze conventional retrieval. Where pure text retrieval hits a hard ceiling between 10% and 23% accuracy (scoring literally 0% on all multi-paper lists, maxima, and counts), uniting text retrieval with structured SQL execution more than doubles performance to 51.5% on historical and 50.8% on prospective literature—a net improvement of $33.0 \pm 1.8$ percentage points over hybrid text search. Multi-document aggregation queries surge from 0% to 40%–67% accuracy, and single-paper lookups jump to 70%–80% as the agent combines paper titles with direct tabular lookups. Crucially, the accuracy advantage persists identically on literature published after the model's pretraining cutoff ($31.0 \pm 2.5$ points), proving that structured execution operates as a genuine reasoning and retrieval mechanism rather than a surrogate for memorized training data.

Ultimately, these findings establish a foundational principle for scientific literature interfaces: experimental research literature functions essentially as an unstructured database of measurements. Because quantitative scientific inquiries demand relational operations, automated extraction of device records into structured relational schemas must be treated as a primary foundation for literature QA, rather than as an independent downstream curation effort.

## 2 Benchmark and question-answering systems

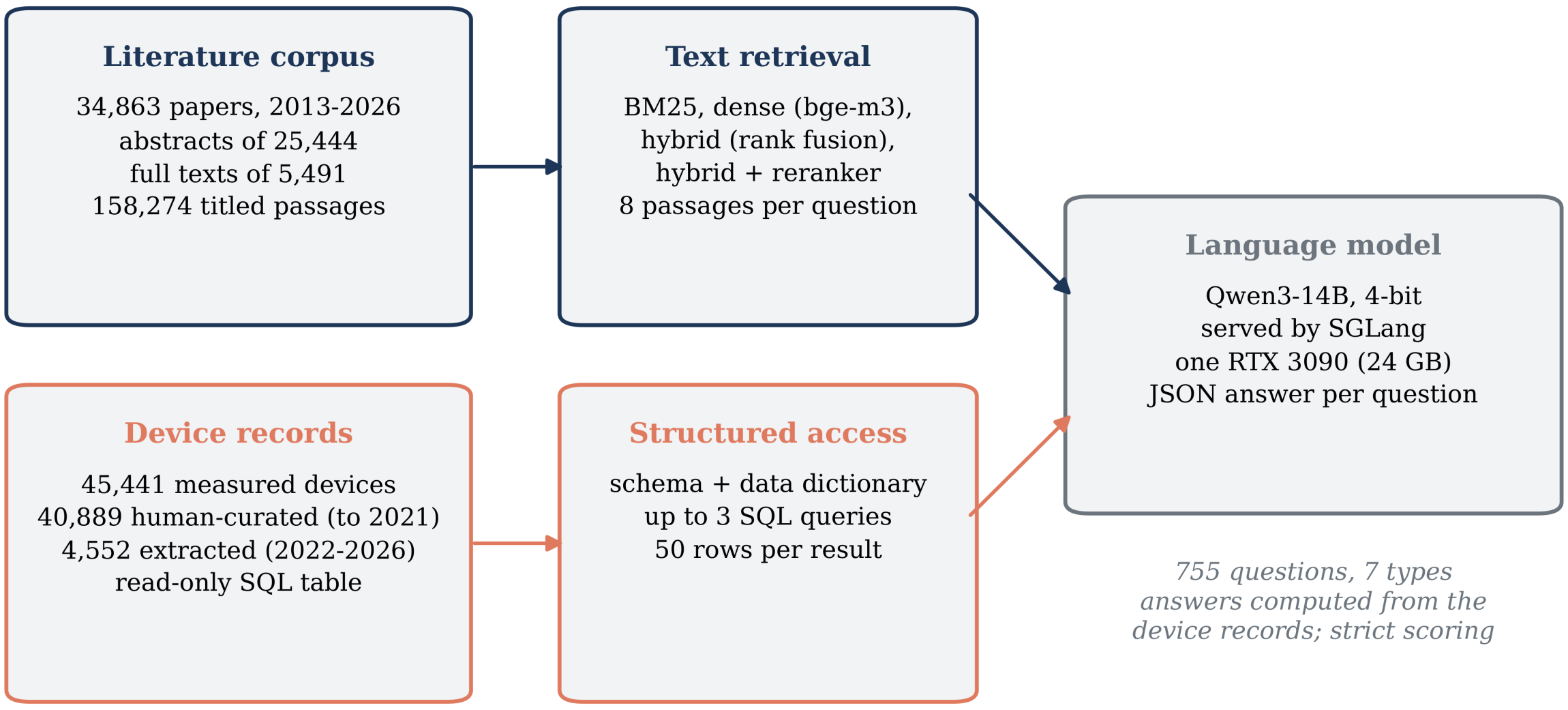

Figure 1: **Benchmark and question-answering systems.** The literature corpus is indexed as titled passages for text retrieval, and the measured device records form a read-only SQL table. Every system uses the same language model, served on one consumer GPU, and differs only in the context it receives: nothing, retrieved passages, or retrieved passages together with SQL access to the device records. The 755 benchmark questions are generated from the device records, so each answer is computed rather than written.

### 2.1 Literature corpus and device records

The evaluation corpus comprises 34,863 perovskite solar cell publications (Fig. 1): 7,352 DOIs indexed in the community Perovskite Database [6] and resolved in OpenAlex, alongside 27,511 papers published from 2022 onward. Abstract metadata were harvested via OpenAlex, Semantic Scholar, and Crossref, while full-text articles were acquired lawfully through open-access licenses, publisher text-and-data-mining interfaces, and repository copies (yielding 25,444 abstracts and 5,491 full-text documents processed locally without redistribution). Full texts were segmented by section headers, stripped of bibliography and back-matter, and chunked into 300-word passages with 50-word overlaps, yielding 158,274 passages across 25,662 papers. Every full-text chunk was prefixed with the paper title and section name, and abstracts were indexed alongside their titles.

For the 2013–2021 historical corpus, device metrics were drawn directly from the human-annotated Perovskite Database [6], filtering for power conversion efficiencies between 0.1% and 30% (40,889 devices across 7,269 papers). For 2022–2026 literature, records were extracted from full texts using an LLM-guided extraction pipeline governed by an explicit schema manual (4,552 devices from 1,456 papers). To guarantee physical integrity, records were retained only when open-circuit voltage ($V_{oc}$), short-circuit current density ($J_{sc}$), fill factor (FF), and efficiency were mutually self-consistent (the product $V_{oc} \times J_{sc} \times$ FF within 5% of recorded efficiency), conformed to standard 1-sun test conditions, and contained no duplicate records across documents. The human-curated historical corpus serves as an established, noise-free reference for validating findings on the prospective extracted subset.

### 2.2 Questions with computed answers

Devices were organized into categorical groups based on recorded metadata: ten composition families spanning A-, B-, and X-site chemistries (e.g., $MAPbI_3$, mixed FA/Cs lead halides, and tin–lead alloys), seven hole-transport and four electron-transport material classes, device polarity (n-i-p vs. p-i-n), and two-year publication cohorts. Using programmatic templates, we generated 755 questions spanning seven distinct inquiry classes: 371 historical and 384 recent queries (Table 1). While single-paper lookups ask for the peak reported efficiency of a specifically titled paper, the remaining six types require synthesizing all relevant devices in the database.

To eliminate ambiguous or ill-posed queries, candidate questions had to satisfy strict statistical criteria: medians required at least 15 devices across ≥5 distinct papers; comparisons and trends required ≥10 devices per group and a margin ≥0.5 percentage points; and list queries required between 2 and 6 papers separated from unselected papers by at least 0.3 percentage points. Questions were stratified and sampled with fixed pseudo-random seeds under identical parameters for both epochs.

Table 1: **Question types of the benchmark.** Counts are for the 2013 to 2021 and 2022 to 2026 periods. Answers are computed from the device records.

| Type | Example | 2013–2021 | 2022–2026 |
|---|---|---|---|
| Lookup | Highest PCE reported in a paper named by its title | 100 | 100 |
| List | Papers reporting a composition and hole layer at or above a PCE | 50 | 50 |
| Maximum | Highest PCE for a composition in a period, and its paper | 33 | 37 |
| Count | Number of papers reporting a composition with an electron layer | 40 | 40 |
| Median | Median PCE, $V_{oc}$, $J_{sc}$ or FF of a device class | 60 | 60 |
| Compare | Which of two hole layers gives the higher median PCE, and by how much | 50 | 50 |
| Trend | Direction and size of the change in median PCE between periods | 38 | 47 |

## 2.3 Retrieval and structured access

All experiments employed Qwen3-14B [9] quantised to 4-bit precision and served via SGLang [10] with greedy decoding (reasoning tokens deactivated), outputting structured JSON according to type-specific schemas. Lexical search used BM25 [11, 12]; dense retrieval used bge-m3 embeddings [13] in an inner-product index; hybrid search merged lexical and dense rankings via reciprocal rank fusion [14]; and reranked setups rescored the top 50 hybrid passages using a bge cross-encoder. Each prompt incorporated eight retrieved passages with full provenance (DOI, year, title, section).

In the structured access variant, the model received the eight hybrid passages alongside the SQLite schema and a concise domain data dictionary mapping chemical nomenclature to database conventions, following established text-to-SQL benchmark designs [15]. The agent was permitted up to three sequential SQL queries (capped at 50 returned rows each) before issuing its final response. All benchmarking was hosted locally on an off-the-shelf workstation equipped with an RTX 3090 GPU (24 GB VRAM).

## 2.4 Scoring

Lookup and maximum queries were marked correct if the numeric value fell within 1% relative error (or ≤0.05 percentage points absolute), with maxima additionally requiring the correct paper citation. Medians required agreement within 5%, counts required exact integer matches, and lists required the exact supporting paper set. For comparisons and trends, both the qualitative direction and the magnitude of the difference (within 25% relative tolerance) were required, preventing false positives from guessing the ubiquitous historical upward trend in efficiency. Retrieval recall represents the fraction of ground-truth supporting papers surfaced in the top-8 passages. Statistical uncertainties reflect 95% bootstrap confidence intervals (10,000 resamples per period), with paired differences evaluated on matching question resamples and reported as estimate ± SE.

# 3 Results and discussion

## 3.1 Performance comparison across retrieval designs

Benchmarking the six system architectures exposes a stark performance hierarchy across retrieval strategies (Fig. 2, Table 2). Baseline generation without retrieval yields near-zero accuracy (1.6% on historical, 6.5% on recent queries), demonstrating that parametric memory alone cannot supply precise photovoltaic parameters. Standard passage retrieval provides modest gains, plateauing between 9.7% and 22.7%, with BM25 lexical matching serving as the strongest single text retriever (16.2% on historical, 22.7% on prospective literature), matching hybrid fusion on earlier papers and surpassing it on recent ones.

Crucially, granting the language model structured SQL execution over the device database sparks a dramatic leap in performance, reaching 51.5% on 2013–2021 literature and 50.8% on 2022–2026 literature. This represents an overall paired advantage of 33.0 ± 1.8 percentage points over hybrid text retrieval (35.0 ± 2.7 historical, 31.0 ± 2.5 prospective) and 31.7 ± 1.9 points over BM25.

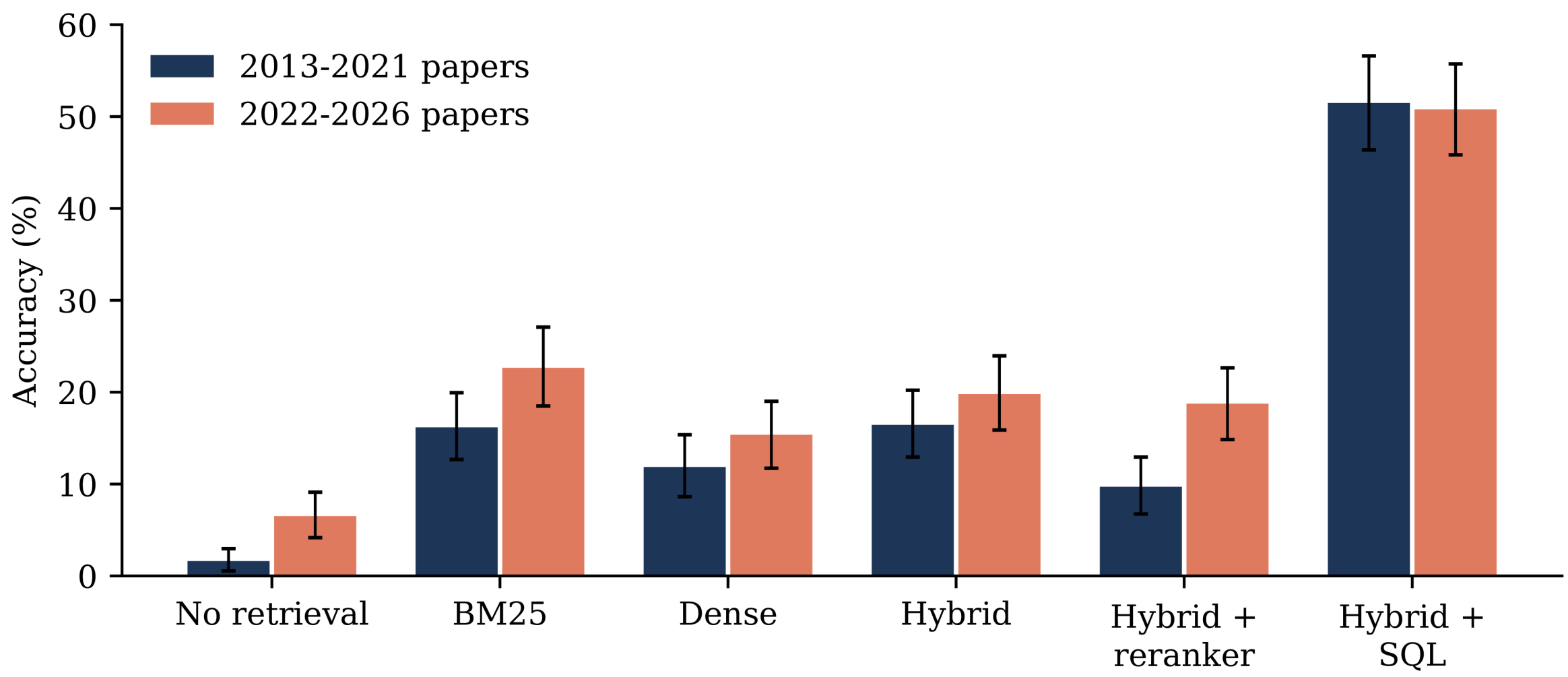

Figure 2: **Accuracy of six question-answering systems on the benchmark.** Bars give the share of questions answered correctly on papers published from 2013 to 2021 (371 questions) and from 2022 to 2026 (384 questions), with 95% bootstrap intervals over questions. All systems use the same language model and differ in the context they receive. Data: Table 2.

## 3.2 Failure modes of text retrieval on multi-document queries

Analyzing performance by question category highlights why pure text retrieval is fundamentally unsuited for quantitative materials informatics (Fig. 3, Table 2). Passage retrieval achieves exactly 0% accuracy on all list, maximum, and count questions across both evaluation periods. Because the evidence supporting these queries is scattered throughout hundreds of distinct documents, retrieving eight 300-word passages achieves less than 5% recall of the required paper set.

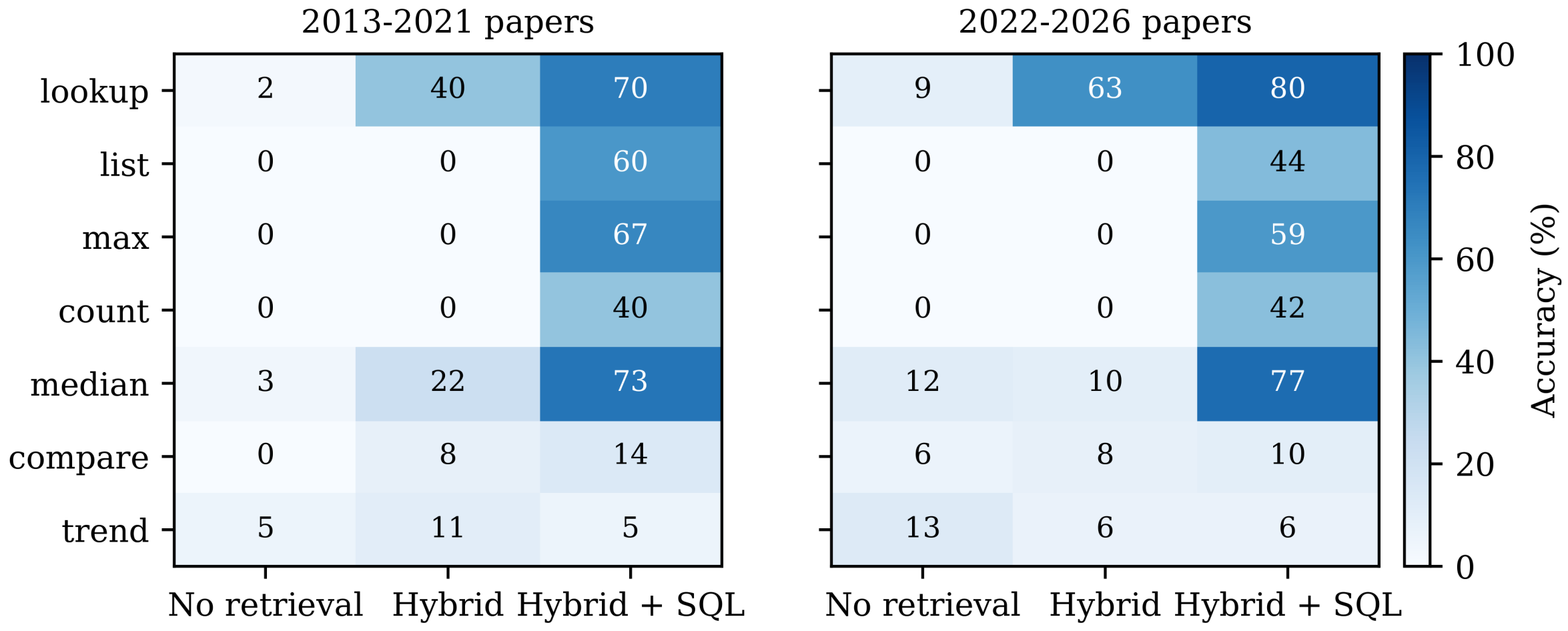

Figure 3: **Accuracy by question type.** Share of questions answered correctly without retrieval, with hybrid text retrieval and with hybrid retrieval plus SQL access to the device table, for papers from 2013 to 2021 (left) and 2022 to 2026 (right). Data: Table 2.

Table 2: **Accuracy (%) by question type and system.** Each cell gives the 2013–2021 / 2022–2026 values. The last two rows give the overall accuracy ± SE for each period (bootstrap over questions).

| Type | No retrieval | BM25 | Dense | Hybrid | Hybrid + reranker | Hybrid + SQL |
|---|---|---|---|---|---|---|
| Lookup | 2 / 9 | 44 / 67 | 24 / 41 | 40 / 63 | 16 / 42 | 70 / 80 |
| List | 0 / 0 | 0 / 0 | 0 / 0 | 0 / 0 | 0 / 0 | 60 / 44 |
| Maximum | 0 / 0 | 0 / 0 | 0 / 0 | 0 / 0 | 0 / 0 | 67 / 59 |
| Count | 0 / 0 | 0 / 0 | 0 / 0 | 0 / 0 | 0 / 0 | 40 / 42 |
| Median | 3 / 12 | 17 / 22 | 17 / 18 | 22 / 10 | 20 / 30 | 73 / 77 |
| Compare | 0 / 6 | 10 / 10 | 8 / 6 | 8 / 8 | 10 / 12 | 14 / 10 |
| Trend | 5 / 13 | 3 / 4 | 16 / 9 | 11 / 6 | 8 / 13 | 5 / 6 |
| Overall, 2013–2021 | 1.6 ± 0.6 | 16.2 ± 1.9 | 11.9 ± 1.7 | 16.4 ± 1.9 | 9.7 ± 1.6 | 51.5 ± 2.6 |
| Overall, 2022–2026 | 6.5 ± 1.3 | 22.7 ± 2.2 | 15.4 ± 1.9 | 19.8 ± 2.1 | 18.8 ± 2.0 | 50.8 ± 2.5 |

Under SQL access, the same 14B model resolves 60% and 44% of list tasks, 67% and 59% of maxima, and 40% and 42% of counts. Medians similarly surge from ≤30% to 73% and 77%.

In contrast, comparative and trend queries remain bottlenecks across all architectures (≤16% accuracy). For trend questions, SQL access achieves only 5–6% accuracy—lower than dense retrieval (9–16%) or zero-shot parametric guessing (13%). This limitation arises because SQLite lacks native 'MEDIAN()' aggregation; when prompted to calculate medians across multi-year cohorts, the model substitutes 'AVG()', which is heavily distorted by distribution skew and fails the 25% tolerance threshold.

### 3.3 Single-paper lookups, reranker pathology, and title prefixing

Single-paper lookups directly isolate document retrieval mechanics because the target study is cited by title in the prompt. On these queries, text retrieval recall strongly predicts performance: BM25 and hybrid search place a passage from the target paper in the top-8 in 94–99% of queries, enabling 40–67% lookup accuracy, whereas dense retrieval achieves only 62–68% recall (24–41% accuracy). Structured SQL access elevates lookup accuracy further to 70% (historical) and 80% (prospective) by directly querying device records by paper title, bypassing extraction errors from text snippets.

Rescoring candidates with a cross-encoder reranker collapses recall to 49% and 65%, reducing overall lookup accuracy by 3.8 ± 1.4 percentage points (and −6.7 ± 1.9 points on his-torical literature). Because lookup prompts ask for highest reported efficiencies, the cross-encoder prioritizes generic passages discussing record-setting efficiencies from unrelated papers, evicting the specifically cited paper from the context window.

Conversely, prepending paper titles and section tags to every text chunk significantly sharpens dense retrieval, boosting lookup recall from 42% to 62% on older literature and generating net accuracy improvements of $+2.0 \pm 0.9$ points for BM25 and $+2.6 \pm 1.1$ points for dense retrieval.

### 3.4 Evaluating memorization across publication epochs

Testing models across distinct publication epochs reveals that these accuracy gains reflect genuine system capabilities rather than training data memorization. In the zero-shot baseline, accuracy remains low regardless of era (1.6% vs. 6.5%), with the small boost on recent papers attributable to educated guessing within contemporary efficiency distributions.

Under structured SQL access, system accuracy is virtually indistinguishable between 2013–2021 literature (51.5%) and 2022–2026 literature (50.8%), showing consistent paired gains over hybrid search of $35.0 \pm 2.7$ and $31.0 \pm 2.5$ percentage points. The effectiveness of structured query interfaces is thus invariant to paper age, holding just as strongly on literature released after model pretraining.

### 3.5 Execution budget and hardware feasibility

All experimental evaluations were conducted locally on an off-the-shelf RTX 3090 GPU (24 GB VRAM). Operating under 4-bit quantisation via SGLang, Qwen3-14B consumed 19.8 GB of memory, leaving sufficient headroom to load the embedding and cross-encoder models concurrently.

Generating dense embeddings across 158,274 passages required 16 minutes. Answering all 755 questions took approximately 27 minutes for passage retrieval and 70 minutes under multi-turn SQL generation. Evaluating the six primary system architectures alongside the three title-prefixing ablations ran in under five hours on local hardware without cloud services, ensuring complete privacy for proprietary literature corpora.

## 4 Conclusion

This work introduces a quantitative question-answering benchmark for perovskite solar cells evaluated against thousands of verified experimental device records, systematically contrasting passage retrieval with structured table access.

Our findings indicate that passage-based RAG architectures possess an inherent ceiling when applied to quantitative materials literature: while capable of surfacing single-paper metrics, they fail completely on questions requiring multi-document aggregation. Granting a compact 14B model read-only SQL access over an extracted device table more than doubles accuracy across both historical and prospective literature cohorts, successfully answering complex aggregations that text retrieval cannot resolve.

These results demonstrate that literature-scale intelligence in the experimental sciences depends foremost on data representation. Transforming scientific literature from flat prose into structured measurement tables is not merely a curation byproduct, but the foundational architecture required for reliable scientific discovery assistants.

**Data availability.** The question set, the computed answers and the list of DOIs are available from the authors on reasonable request. Full texts are not redistributed.

**Acknowledgements.** The authors used the large language models Claude (Anthropic) and Gemini (Google) as assistants: Claude in extracting the 2022 to 2026 device records, in drafting parts of the manuscript and in writing the analysis code, and Gemini in drafting the manuscript. All text, code and results were reviewed, verified and revised by the authors, who take full responsibility for the content of this article.

**Competing interests.** The authors declare no competing interests.

## References

[1] John Dagdelen, Alexander Dunn, Sanghoon Lee, et al. Structured information extraction from scientific text with large language models. *Nature Communications*, 15:1418, 2024.

[2] Maciej P. Polak and Dane Morgan. Extracting accurate materials data from research papers with conversational language models and prompt engineering. *Nature Communications*, 15:1569, 2024.

[3] Jakub Lála, Odhran O'Donoghue, Aleksandar Shtedritski, et al. PaperQA: Retrieval-augmented generative agent for scientific research. *arXiv preprint arXiv:2312.07559*, 2023.

[4] Patrick Lewis, Ethan Perez, Aleksandra Piktus, et al. Retrieval-augmented generation for knowledge-intensive NLP tasks. In Advances in Neural Information Processing Systems, 2020.

[5] Michael D. Skarlinski, Sam Cox, Jon M. Laurent, et al. Language agents achieve superhuman synthesis of scientific knowledge. *arXiv preprint arXiv:2409.13740*, 2024.

[6] T. Jesper Jacobsson, Adam Hultqvist, Alberto García-Fernández, et al. An open-access database and analysis tool for perovskite solar cells based on the FAIR data principles. *Nature Energy*, 7:107–115, 2022.

[7] Xiang Liu, Penglei Sun, et al. Perovskite-LLM: Knowledge-enhanced large language models for perovskite solar cell research. In Findings of the Association for Computational Linguistics: EMNLP 2025, 2025. arXiv:2502.12669.

[8] Asim Biswal, Liana Patel, Siddarth Jha, et al. Text2SQL is not enough: Unifying AI and databases with TAG. In Conference on Innovative Data Systems Research (CIDR), 2025. arXiv:2408.14717.

[9] Qwen Team. Qwen3 technical report. *arXiv preprint arXiv:2505.09388*, 2025.

[10] Lianmin Zheng, Liangsheng Yin, Zhiqiang Xie, et al. SGLang: Efficient execution of structured language model programs. In Advances in Neural Information Processing Systems, 2024. arXiv:2312.07104.

[11] Stephen Robertson and Hugo Zaragoza. The probabilistic relevance framework: BM25 and beyond. *Foundations and Trends in Information Retrieval*, 3(4):333–389, 2009.

[12] Xing Han Lù. BM25S: Orders of magnitude faster lexical search via eager sparse scoring. *arXiv preprint arXiv:2407.03618*, 2024.

[13] Jianlv Chen, Shitao Xiao, Peitian Zhang, et al. BGE M3-Embedding: Multi-lingual, multi-functionality, multi-granularity text embeddings through self-knowledge distillation. *arXiv preprint arXiv:2402.03216*, 2024.

[14] Gordon V. Cormack, Charles L. A. Clarke, and Stefan Büttcher. Reciprocal rank fusion outperforms Condorcet and individual rank learning methods. In Proceedings of the 32nd International ACM SIGIR Conference, pages 758–759, 2009.

[15] Jinyang Li, Binyuan Hui, Ge Qu, et al. Can LLM already serve as a database interface? A BIg bench for large-scale database grounded text-to-SQLs. In Advances in Neural Information Processing Systems (Datasets and Benchmarks), 2023.